\documentclass[runningheads]{llncs}
\usepackage[T1]{fontenc}
\usepackage[pagebackref,breaklinks,colorlinks]{hyperref}
\usepackage{lettrine}
\usepackage[ruled,vlined]{algorithm2e}
\usepackage{multirow}
\usepackage{booktabs,tabularx,makecell}
\usepackage{pifont} 
\usepackage{array,tabularx,booktabs,rotating}
\newcolumntype{Y}{>{\centering\arraybackslash}X} 
\usepackage{colortbl}
\usepackage{amsmath}
\usepackage{amssymb}
\usepackage{cleveref}
\usepackage{xcolor}
\usepackage{float}
\usepackage{pifont}
\usepackage{cleveref}
\Crefname{figure}{Fig.}{Figs.}
\crefname{figure}{Fig.}{Figs.}
\newcommand{\cmark}{\ding{51}}
\newcommand{\xmark}{\ding{55}}

\usepackage{graphicx,verbatim}
\begin{document}
%
\title{Structure-Adaptive Sparse Diffusion in Voxel Space for 3D Medical Image Enhancement}



%

\author{Anonymized Authors}  
\authorrunning{Anonymized Author et al.}
\institute{Anonymized Affiliations \\
    \email{email@anonymized.com}}

\author{Hongxu Jiang\inst{1} \and
Fei Li\inst{2} \and
Boxiao Yu\inst{3} \and
Ying Zhang\inst{4} \and
Kaleb Smith\inst{5} \and
Kuang~Gong\inst{3} \and
Wei Shao\inst{2}\thanks{Corresponding author: Wei Shao (e-mail: weishao@ufl.edu).}}

\authorrunning{Jiang et al.}

\institute{
Department of Electrical and Computer Engineering, University of Florida, Gainesville, FL, USA\\
\email{hongxu.jiang@ufl.edu} \and
Department of Medicine, University of Florida, Gainesville, FL, USA\\
\email{fei.Li@medicine.ufl.edu, weishao@ufl.edu} \and
Department of Biomedical Engineering, University of Florida, Gainesville, FL, USA\\
\email{boxiao.yu@ufl.edu, kgong@bme.ufl.edu} \and
Research Computing, University of Florida, Gainesville, FL, USA\\
\email{yingz@ufl.edu} \and
NVIDIA, Santa Clara, CA, USA\\
\email{kasmith@nvidia.com}
}
  
\maketitle              
\begin{abstract}
Three-dimensional (3D) medical image enhancement, including denoising and super-resolution, is critical for clinical diagnosis in CT, PET, and MRI. Although diffusion models have shown remarkable success in 2D medical imaging, scaling them to high-resolution 3D volumes remains computationally prohibitive due to lengthy diffusion trajectories over high-dimensional volumetric data. We observe that in conditional enhancement, strong anatomical priors in the degraded input render dense noise schedules largely redundant. Leveraging this insight, we propose a sparse voxel-space diffusion framework that trains and samples on a compact set of uniformly subsampled timesteps. The network predicts clean data directly on the data manifold, supervised in velocity space for stable gradient scaling. A lightweight Structure-aware Trajectory Modulation (STM) module recalibrates time embeddings at each network block based on local anatomical content, enabling structure-adaptive denoising over the shared sparse schedule. Operating directly in voxel space, our framework preserves fine anatomical detail without lossy compression while achieving up to $10\times$ training acceleration. Experiments on four datasets spanning CT, PET, and MRI demonstrate state-of-the-art performance on both denoising and super-resolution tasks. Our code is publicly available at: \url{https://github.com/mirthAI/sparse-3d-diffusion}.


\end{abstract}

\section{Introduction}
\label{sec:intro}
Three-dimensional (3D) medical imaging modalities, such as computed tomography (CT), magnetic resonance imaging (MRI), and positron emission tomography (PET), are central to modern clinical diagnosis and treatment planning~\cite{hussain2022modern}. In practice, however, image acquisition is often constrained by radiation dose, scan duration, and hardware limitations, resulting in low signal-to-noise ratio and limited spatial resolution. These degradations obscure subtle anatomical structures and compromise diagnostic confidence, motivating extensive research into 3D image enhancement tasks, including denoising ~\cite{yu2024pet,gao2023corediff} and super-resolution ~\cite{liu2025diffusion,zhang2025geodesic}.
Early enhancement approaches operated in a slice-by-slice or 2.5D manner, often resulting in through-plane inconsistencies due to limited volumetric context. Fully 3D convolutional neural networks~\cite{chaudhari2018super,pham2019multiscale} and generative adversarial networks~\cite{sanchez2018brain,zhang2022soup} were later introduced, but convolutional models often produce over-smooth details and adversarial methods remain difficult to optimize~\cite{sanchez2018brain,uzunova2020memory}. 


Diffusion probabilistic models~\cite{ho2020denoising,song2020denoising,song2020score} have recently emerged as a powerful alternative offering more stable training, better distribution coverage, and improved perceptual fidelity~\cite{khader2023denoising,yu2024pet}. To reduce computational costs, most existing methods operate diffusion in compressed representations, including VAE-learned latents~\cite{zhu2023make,khader2023denoising}, wavelet decompositions~~\cite{friedrich2024wdm,zheng20253d}, or downsampled inputs~\cite{bieder2024memory,dorjsembe2024conditional}. While efficient, such compressions inevitably discard high-frequency anatomical details (e.g., cortical folds and lung fissures), imposing representational bottlenecks that fundamentally limit enhancement quality. A natural alternative is returning to the original voxel space, thereby preserving full spatial resolution and avoiding information loss. However, existing diffusion models learn to predict noise or velocity targets that reside in complex, high-dimensional spaces far from the data manifold~\cite{li2025back,ma2026pixelgen}, requiring both prolonged training to capture such mappings and hundreds of iterative denoising steps to recover a clean volume, making direct 3D voxel-space generation prohibitively expensive.

To mitigate this overhead, we revisit the problem from a conditional modeling perspective. In enhancement tasks, the degraded input already encodes strong anatomical priors, fundamentally reducing the learning complexity compared to unconditional generation. The key is to choose a prediction target that preserves this advantage: predicting noise or velocity maps the network output away from the data manifold, discarding the structural proximity between the degraded input and the clean target. We instead adopt clean data $x_0$ prediction, which keeps the network output on the data manifold throughout training, allowing the model to to directly refine the conditioning prior rather than learn a detour through high-variance noise or velocity spaces.

This closer alignment between prediction target and enhancement objective yields smoother denoising behaviors in which neighboring timesteps produce highly similar outputs~\cite{hang2023efficient,choi2022perception,go2023addressing,ma2025decouple}, naturally motivating sparse diffusion on a subset of representative timesteps. While~\cite{jiang2025fast} showed that a fixed sparse schedule suffices for 2D conditional generation, extending it to patch-based 3D diffusion reveals a further challenge: high-resolution patches across diverse anatomical regions (e.g., homogeneous liver tissue versus intricate vascular structures) exhibit varied denoising dynamics that a single global schedule cannot accommodate.

To this end, we propose a structure-aware sparse diffusion framework that operates directly in 3D voxel space for medical image enhancement. The framework combines three complementary designs: $x_0$-prediction with velocity supervision for stable on-manifold learning, sparse diffusion that eliminates redundant denoising steps, and a Structure-aware Trajectory Modulation (STM) module that recalibrates the time embedding according to local anatomical complexity, enabling structure-adaptive denoising at each step. Together, these components achieve up to 10$\times$ faster training while delivering state-of-the-art performance across CT, PET, and MRI denoising and super-resolution benchmarks.

The main contributions of this work are as follows:
\vspace{-0.1cm}
\begin{enumerate}
    \item We propose a sparse diffusion framework in 3D voxel space that combines $x_0$-prediction with velocity supervision and timestep subsampling, achieving up to $10\times$ faster training while improving enhancement quality.
    \item We introduce Structure-aware Trajectory Modulation (STM), which leverages tri-planar structural encoding and block-wise time embedding modulation to adapt denoising trajectories to local anatomical structure.
    \item Extensive experiments on four datasets spanning CT, PET, and MRI deliver state-of-the-art performance on both denoising and super-resolution tasks.
\end{enumerate}

\section{Methodology}
\vspace{-0.3cm}
\subsection{Overview}
\begin{figure*}[!hbt]
\includegraphics[width=\linewidth]{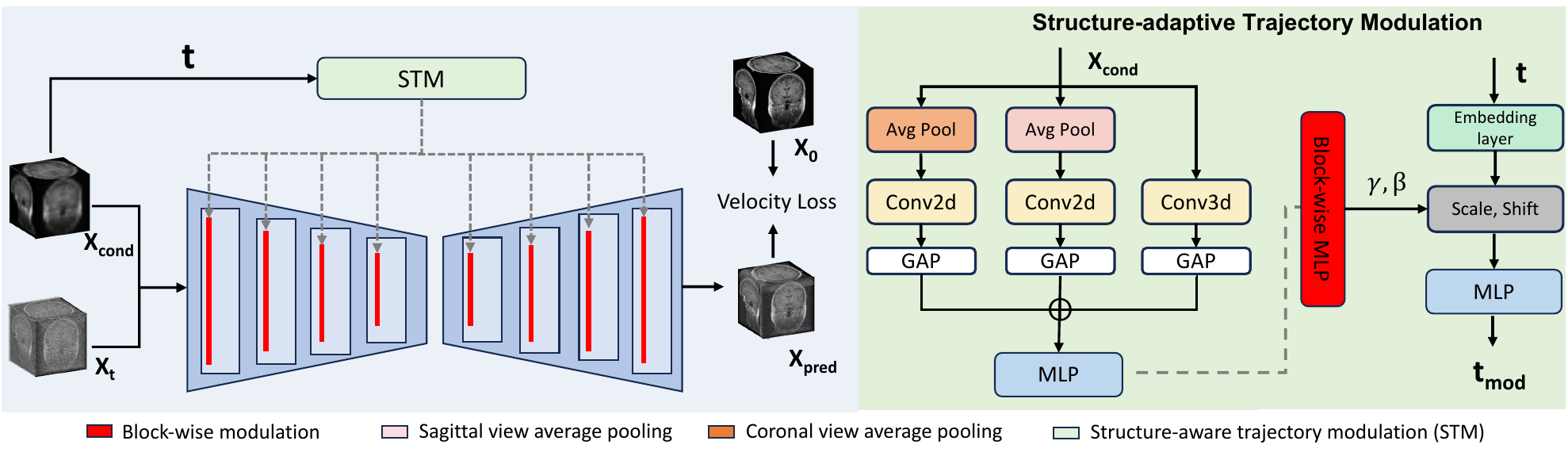}
\centering
\vspace{-0.7cm}
\caption{
\textbf{Overview of the proposed framework.}
}
\label{fig:framework}
\end{figure*}

The proposed framework, illustrated in \cref{fig:framework}, performs structure-adaptive sparse diffusion directly in voxel space for 3D medical image enhancement. High-resolution volumes are first partitioned into overlapping 3D patches to enable memory-efficient processing while preserving local spatial context. During diffusion, the model takes the noisy patch $x_t$ and degraded input $x_{\text{cond}}$ as joint inputs and predicts the clean target $x_0$, supervised by a velocity loss for stable on-manifold learning. Training operates on a subset of uniformly subsampled timesteps, forming a sparse trajectory that eliminates redundant denoising steps. A Structure-aware Trajectory Modulation (STM) module further extracts structural cues from $x_{\text{cond}}$ to recalibrate time embeddings at each UNet block, adapting the denoising behavior to local anatomical complexity.

\vspace{-0.3cm}
\vspace{-0.7cm}

\subsection{Sparse Voxel-Space Diffusion}
Let $x_0 \in \mathbb{R}^{D \times H \times W}$ denote a high-quality 3D medical volume and $y \in \mathbb{R}^{D \times H \times W}$ its degraded input. To enable tractable processing under GPU memory constraints, we extract overlapping 3D patches $\{x_0^{(i)}, y^{(i)}\}_{i=1}^{N}$ of size $d \times h \times w$ from the volume pair.
The forward diffusion process corrupts $x_0$ via $x_t = \sqrt{\bar{\alpha}_t}\, x_0 + \sqrt{1 - \bar{\alpha}_t}\, \epsilon$, where $\bar{\alpha}_t = \prod_{s=1}^{t}(1 - \beta_s)$ and $\epsilon \sim \mathcal{N}(0, \mathbf{I})$.

\vspace{0.8mm}

\noindent\textbf{\boldmath$x_0$-Prediction.}
In conditional enhancement, the degraded input $y$ already provides strong anatomical priors, restricting the problem to local refinement rather than unconditioned generation. Rather than regressing noise $\epsilon$ or velocity $v$, which reside in high-variance spaces off the data manifold, we let the denoising network $f_\theta$ directly predict the clean data: $\hat{x}_0 = f_\theta(x_t, y, t)$.

\noindent\textbf{Sparse Timestep Scheduling.}
The $x_0$-prediction formulation yields smoother denoising trajectories with high inter-step redundancy, as neighboring noise levels produce highly similar outputs. We exploit this by uniformly subsampling $K$ timesteps from the full schedule of $T$ steps (e.g., $T$=1000) to form a sparse schedule $\mathcal{S} = \{s_1, \dots, s_K\}$ where $K \ll T$. (we use $K{=}5$; see ablation in \cref{tab:ablation_timesteps}). Both forward and reverse processes then operate exclusively on $\mathcal{S}$, whereas conventional models require the full $T$ steps.

\noindent\textbf{Velocity Supervision.}
While $x_0$-prediction keeps the network output on the data manifold, directly supervising in $x_0$-space leads to unbalanced gradient magnitudes across timesteps: the loss scale vanishes near $t{=}0$ and explodes near $t{=}T$~\cite{salimans2022progressive,ho2020denoising}. We therefore reformulate the supervision in velocity space. Substituting $\epsilon = (x_t - \sqrt{\bar{\alpha}_t}\, x_0) / \sqrt{1 - \bar{\alpha}_t}$ into the velocity definition $v_t = \sqrt{\bar{\alpha}_t}\, \epsilon - \sqrt{1 - \bar{\alpha}_t}\, x_0$ yields $v_t = (\sqrt{\bar{\alpha}_t}\, x_t - x_0) / \sqrt{1 - \bar{\alpha}_t}$. We apply the same conversion to the network prediction $\hat{x}_0$ to obtain:
\begin{equation}
    \hat{v}_t = \frac{\sqrt{\bar{\alpha}_t}\, x_t - \hat{x}_0}
    {\sqrt{1 - \bar{\alpha}_t}}.
    \label{eq:vel_convert}
\end{equation}
This retains the benefit of on-manifold prediction while 
producing uniformly scaled gradients across all timesteps in 
$\mathcal{S}$, stabilizing training under our sparse schedule. 
The training objective is given by:
\begin{equation}
    \mathcal{L}_{\text{base}} = \mathbb{E}_{x_0,\, t \sim 
    \mathcal{U}(\mathcal{S}),\, \epsilon \sim \mathcal{N}(0, 
    \mathbf{I})} \left[\left\| v_t - \hat{v}_t \right\|_2^2 \right].
    \label{eq:loss_base}
\end{equation}

\vspace{-0.3cm}

\subsection{Structure-aware Trajectory Modulation}
The sparse diffusion process described above applies a globally uniform denoising schedule across all patches regardless of their content. In practice, however, 3D medical volumes exhibit pronounced anatomical heterogeneity: patches containing fine structural detail (e.g., bronchial trees, cortical folds) require more gradual denoising, whereas homogeneous regions (e.g., soft tissue, background) can tolerate coarser adjustments. To address this, we propose Structure-aware Trajectory Modulation (STM), which dynamically conditions the denoising process on local anatomical content, enabling structure-adaptive denoising behavior across the shared sparse schedule.

\noindent\textbf{Multi-structure Encoder.}
To capture structural context from complementary spatial dimensions, we design a lightweight multi-structure encoder $E_\phi$ as shown in \cref{fig:framework}. A volumetric branch extracts holistic 3D structural features from the conditioning input $x_{\text{cond}}$ through 3D convolutions followed by global average pooling (GAP). In parallel, two planar branches operate on mean projections of $x_{\text{cond}}$ along the coronal and sagittal axes, each processed by 2D convolutions and GAP to capture orientation-specific 2D structural patterns that may be diluted in purely volumetric pooling. The three branch outputs are concatenated and fused through an MLP to produce a compact structural representation:
\begin{equation}
    \mathbf{c} = \mathrm{MLP}\!\left(\left[E_\phi^{\mathrm{3D}}(x_{\text{cond}});\; E_\phi^{\mathrm{cor}}(\tilde{x}_{\text{cond}}^{\mathrm{cor}});\; E_\phi^{\mathrm{sag}}(\tilde{x}_{\text{cond}}^{\mathrm{sag}})\right]\right) \in \mathbb{R}^{d_c},
\end{equation}
where $\tilde{x}_{\text{cond}}^{\mathrm{cor}}$ and $\tilde{x}_{\text{cond}}^{\mathrm{sag}}$ denote mean projections along the anterior--posterior and lateral axes, respectively.

\noindent\textbf{Block-wise Trajectory Modulation.}
The structural representation $\mathbf{c}$ modulates the denoising process by recalibrating the time embedding at every residual block of the UNet. Each block contains a dedicated MLP that maps $\mathbf{c}$ to block-specific affine parameters:
\begin{equation}
    (\gamma_k, \beta_k) = \mathrm{split}\!\left(\mathrm{MLP}_k(\mathbf{c})\right),
\end{equation}
which then modulate the time embedding $\mathbf{e}_t$ to produce a structure-conditioned variant:
\begin{equation}
    \mathbf{t}_{\text{mod}}^{(k)} = (1 + \gamma_k) \odot \mathbf{e}_t + \beta_k,
\end{equation}
where $\mathbf{e}_t$ is obtained from timestep $t$ via a sinusoidal embedding layer and $k$ indexes the residual blocks of the UNet. The block-wise design allows each network depth to learn its own structure-time interaction, as shallow blocks may prioritize coarse anatomical layout while deeper blocks attend to fine-grained detail. Since the time embedding governs how aggressively the network denoises at each step, modulating it with structural information effectively reshapes the denoising trajectory: the same physical timestep induces different denoising strengths depending on local anatomical complexity, enabling structurally distinct patches to follow individualized denoising paths over the shared sparse schedule.

\vspace{-0.3cm}
\section{Experimental Results}

\textbf{Datasets.}
We evaluate the proposed approach on two 3D medical image enhancement tasks: image denoising and image super-resolution, using four publicly available datasets that span CT, PET, and MRI modalities. For CT image denoising, we use the lung LDCT-and-Projection-data dataset \cite{moen2021low} with image sizes ranging from 512$\times$512$\times$300 to 512$\times$512$\times$400. We used 50 patients in total, with 40 for training and 10 for testing. For PET image denoising, we use brain FDG PET dataset \cite{gong2024pet} with an image size of 256$\times$256$\times$89. We used 70 patients for training and 18 for evaluation. For CTA image super-resolution, the AortaSeg24 dataset \cite{imran2025multi} were used with image sizes ranging from 350$\times$350$\times$500 to 520$\times$520$\times$800. We used 50 patients for training and 10 for evaluation. For MRI super-resolution, we use UHB FCD Lesion Brain MRI Dataset \cite{schuch2023open} with image sizes of 320$\times$320$\times$208. We used 120 scans in total, with 96 for training and 24 for testing. All volumes were normalized to [-1, 1].

\noindent\textbf{Model Evaluation.}
We evaluate enhancement quality using four 3D metrics: PSNR, SSIM, MS-SSIM, and HFEN. We compare against four representative 3D diffusion baselines: 3D LDM \cite{khader2023denoising} (latent-space), 3D WDM \cite{friedrich2024wdm} (wavelet-domain), 3D DDPM \cite{yu2024pet}, and 3D DDIM \cite{song2020denoising}, covering the dominant paradigms of 3D medical image diffusion.

\noindent\textbf{Quantitative Results.}
\Cref{tab:quantitative} presents quantitative results across all datasets and tasks. Our method consistently achieves the best or second-best scores on every metric. On lung CT denoising, our model improves PSNR by +0.57\,dB and SSIM by +0.016 over the strongest baseline, while on aorta CTA it surpasses 3D DDIM by +0.46\,dB PSNR with uniformly better SSIM, MS-SSIM, and HFEN, indicating more faithful recovery of fine vascular structures. A consistent improvement is observed across all remaining datasets as well. We attribute these gains to the complementary strengths of the framework: operating in voxel space avoids the representational loss of latent compression, sparse scheduling eliminates redundant steps that cause repeated smoothing, and STM adapts the denoising behavior to local anatomical complexity, preserving fine details in structurally rich regions while allowing more aggressive denoising elsewhere.

\begin{table}[!hbt]
\centering
\caption{Quantitative comparison on denoising and $4\times$ super-resolution tasks across four datasets. Best results are in \textbf{\textcolor{red}{red}}, second-best are \textcolor{blue}{\underline{blue}}.}
\vspace{-0.1cm}
\label{tab:quantitative}
\renewcommand{\arraystretch}{1.08}
\scriptsize
\setlength{\tabcolsep}{2pt}
\begin{tabular}{l|cccc|cccc}
\toprule
\cellcolor{blue!10} & \multicolumn{4}{c|}{\cellcolor{blue!10}\textbf{Lung CT Denoising}} & \multicolumn{4}{c}{\cellcolor{blue!10}\textbf{Brain PET Denoising}} \\
\textbf{Method} & \textbf{PSNR}$\uparrow$ & \textbf{SSIM}$\uparrow$ & \textbf{MS-SSIM}$\uparrow$ & \textbf{HFEN}$\downarrow$ & \textbf{PSNR}$\uparrow$ & \textbf{SSIM}$\uparrow$ & \textbf{MS-SSIM}$\uparrow$ & \textbf{HFEN}$\downarrow$ \\
\midrule
3D LDM  & 28.89 & 0.7380 & 0.9430 & 0.0308 & 35.43 & 0.9605 & 0.9872 & 0.0179 \\
3D WDM  & 29.54 & 0.7578 & 0.9532 & 0.0283 & 36.27 & 0.9624 & 0.9884 & 0.0167 \\
3D DDPM & 29.25 & 0.7414 & 0.9426 & 0.0291 & 36.59 & 0.9621 & 0.9890 & 0.0159 \\
3D DDIM & \textcolor{blue}{\underline{31.25}} & \textcolor{blue}{\underline{0.8073}} & \textcolor{blue}{\underline{0.9613}} & \textcolor{blue}{\underline{0.0230}} & \textcolor{blue}{\underline{36.85}} & \textcolor{blue}{\underline{0.9646}} & \textcolor{blue}{\underline{0.9896}} & \textcolor{blue}{\underline{0.0156}} \\
\rowcolor{gray!15}
\textbf{Ours} & \textbf{\textcolor{red}{31.82}} & \textbf{\textcolor{red}{0.8232}} & \textbf{\textcolor{red}{0.9644}} & \textbf{\textcolor{red}{0.0215}} & \textbf{\textcolor{red}{37.08}} & \textbf{\textcolor{red}{0.9660}} & \textbf{\textcolor{red}{0.9899}} & \textbf{\textcolor{red}{0.0149}} \\
\midrule
\cellcolor{blue!10} & \multicolumn{4}{c|}{\cellcolor{blue!10}\textbf{Aorta CTA Super-Resolution}} & \multicolumn{4}{c}{\cellcolor{blue!10}\textbf{Brain MRI Super-Resolution}} \\
\textbf{Method} & \textbf{PSNR}$\uparrow$ & \textbf{SSIM}$\uparrow$ & \textbf{MS-SSIM}$\uparrow$ & \textbf{HFEN}$\downarrow$ & \textbf{PSNR}$\uparrow$ & \textbf{SSIM}$\uparrow$ & \textbf{MS-SSIM}$\uparrow$ & \textbf{HFEN}$\downarrow$ \\
\midrule
3D LDM  & 40.24 & 0.9765 & 0.9951 & 0.0111 & 33.96 & 0.9613 & 0.9908 & 0.0207 \\
3D WDM  & 40.66 & 0.9772 & 0.9952 & 0.0107 & 34.45 & 0.9645 & 0.9912 & 0.0200 \\
3D DDPM & 40.72 & 0.9742 & 0.9947 & 0.0107 & 33.68 & 0.9592 & 0.9903 & 0.0217 \\
3D DDIM & \textcolor{blue}{\underline{41.52}} & \textcolor{blue}{\underline{0.9798}} & \textcolor{blue}{\underline{0.9957}} & \textcolor{blue}{\underline{0.0096}} & \textcolor{blue}{\underline{35.62}} & \textcolor{blue}{\underline{0.9700}} & \textcolor{blue}{\underline{0.9930}} & \textcolor{blue}{\underline{0.0172}} \\
\rowcolor{gray!15}
\textbf{Ours} & \textbf{\textcolor{red}{41.98}} & \textbf{\textcolor{red}{0.9827}} & \textbf{\textcolor{red}{0.9965}} & \textbf{\textcolor{red}{0.0089}} & \textbf{\textcolor{red}{35.95}} & \textbf{\textcolor{red}{0.9724}} & \textbf{\textcolor{red}{0.9934}} & \textbf{\textcolor{red}{0.0163}} \\
\bottomrule
\end{tabular}
\end{table}

\begin{figure}[!hbt]
\centering
\includegraphics[width=0.9\textwidth]{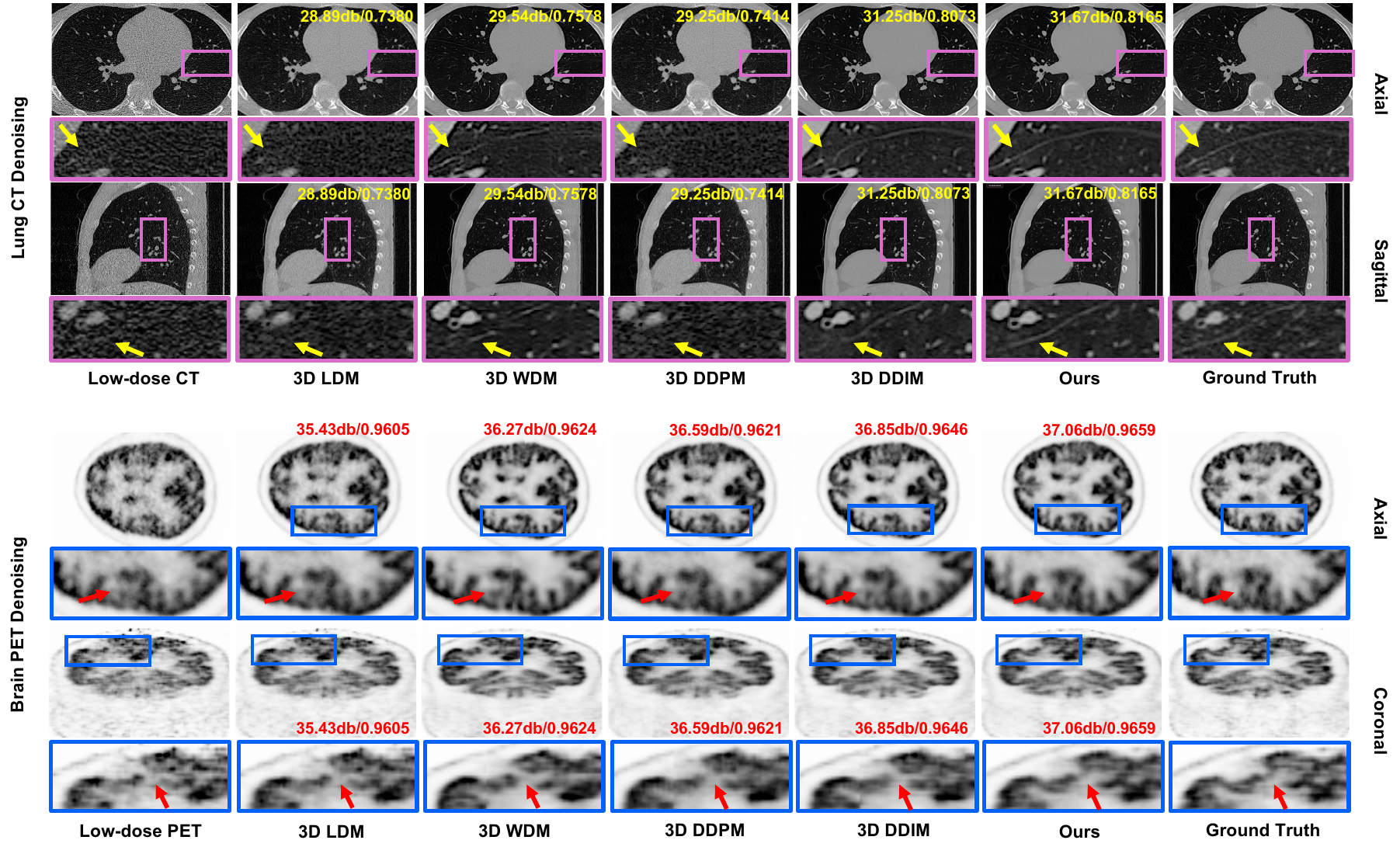}
\vspace{-0.3cm}
\caption{Qualitative denoising results on lung CT and brain PET. Purple and blue boxes highlight lung fissures and subtle anatomical boundaries, respectively. Our model preserves sharper, more structurally consistent details than all baselines.}
\label{fig:denoising_visual}
\end{figure}

\begin{figure}[!hbt]
\centering
\includegraphics[width=0.9\textwidth]{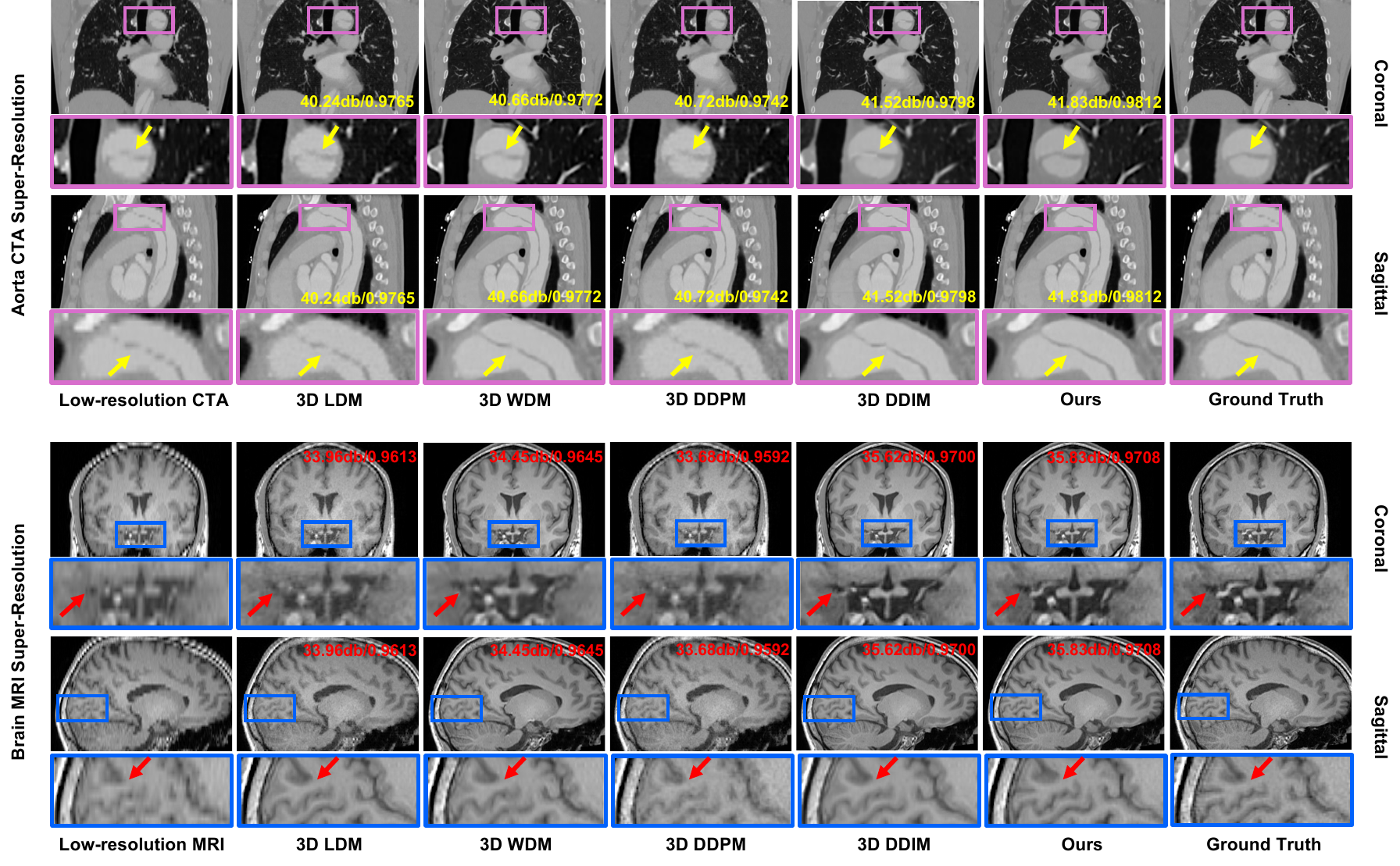}
\vspace{-0.3cm}
\caption{Qualitative $4\times$ super-resolution results on aorta CTA and brain MRI. Highlighted regions show vessel boundaries and cortical structures. Our model recovers finer anatomical details with fewer artifacts than all baselines.}
\label{fig:sr_visual}
\vspace{-0.4cm}
\end{figure}

\vspace{-0.4cm}
\noindent\textbf{Qualitative Results.}
\Cref{fig:denoising_visual} presents denoising comparisons on lung CT and brain PET. In CT, our method reconstructs the sharpest lung fissure lines (purple boxes), closely matching the normal-dose reference, while baselines exhibit blurring or structural loss. In PET, our approach recovers clearer anatomical boundaries (blue boxes) where baselines suffer from oversmoothing.
\Cref{fig:sr_visual} shows $4\times$ super-resolution results on aorta CTA and brain MRI. Our method produces more distinct vessel boundaries in CTA and preserves sharper gray--white matter interfaces in MRI, whereas baselines fail to resolve fine cortical folds and aortic branches. Across both tasks, our proposed method consistently improves fine anatomical detail recovery.

\begin{figure}[!hbt]
\centering
\includegraphics[width=\textwidth]{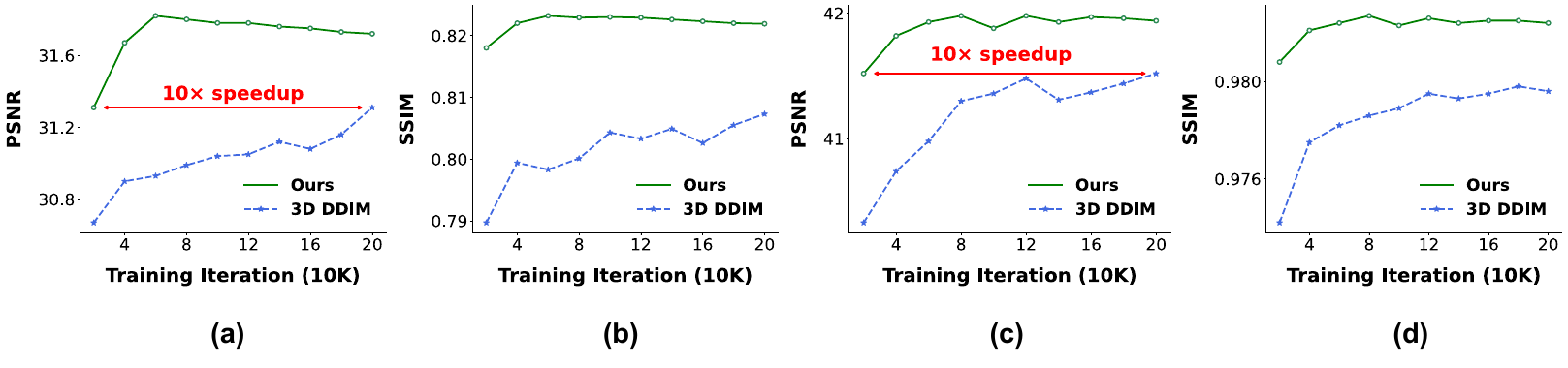}
\vspace{-0.8cm}
\caption{Training convergence curves for lung CT denoising and aorta CTA $4\times$ super-resolution. Our method reaches the baseline's final performance with up to $10\times$ fewer iterations.}
\label{fig:convergence}
\end{figure}

\noindent\textbf{Computational Efficiency.}
We evaluate training efficiency by comparing convergence curves against 3D DDIM on both tasks. As shown in \cref{fig:convergence}, our method achieves up to $10\times$ faster convergence on denoising and super-resolution, reaching higher PSNR at the same iteration count while the baseline remains far from its final performance. This confirms that sparse voxel-space diffusion with STM substantially reduces the computational cost of 3D diffusion without sacrificing enhancement quality.

\vspace{0.3cm}

\noindent\textbf{Ablation Study.}
We first analyze the effect of diffusion step count $K$ on lung CT denoising. As shown in \Cref{tab:ablation_timesteps}, performance peaks at $K{=}5$, with further steps degrading quality due to repeated noise injection--denoising cycles that oversmooth fine structures, confirming that a compact sparse schedule is both sufficient and optimal for conditional enhancement.

\Cref{tab:ablation_components} isolates the contribution of each proposed component. Switching from $\epsilon$ and $v$-prediction to $x_0$-prediction with velocity supervision dramatically improves all metrics. Adding sparse diffusion scheduling improve this quality while reducing training time by $\sim\!10\times$. Finally, incorporating STM yields further gains across all metrics at no additional training cost, validating that structure-adaptive modulation and sparse diffusion are complementary.
\vspace{-0.3cm}

\begin{table}[t]
\centering

\caption{Ablation on diffusion step count $K$ for lung CT denoising (PSNR\,/\,SSIM). Best in \textbf{\textcolor{red}{red}}, second in \textcolor{blue}{\underline{blue}}.}
\label{tab:ablation_timesteps}
\renewcommand{\arraystretch}{1.1}
\scriptsize
\setlength{\tabcolsep}{4pt}
\begin{tabular}{l|ccccc}
\toprule
\rowcolor{blue!10}
\textbf{Method / Step\#} & \textbf{3} & \textbf{5} & \textbf{10} & \textbf{50} & \textbf{100} \\
\midrule
3D DDIM & 30.69 / .7897 & 31.01 / .8020 & 31.25 / .8073 & 30.56 / .7872 & 30.28 / .7839 \\
\rowcolor{gray!15}
\textbf{Ours} & 31.05 / .7968 & \textbf{\textcolor{red}{31.82 / .8232}} & \textcolor{blue}{\underline{31.69 / .8194}} & 31.41 / .8140 & 30.98 / .7988 \\
\bottomrule
\end{tabular}
\vspace{-0.2cm}
\end{table}

\begin{table}[!hbt]
\centering
\caption{Component ablation on lung CT denoising. $x_0$+$v$: $x_0$-prediction with velocity supervision. Sparse: sparse timestep scheduling ($K{=}5$). STM: structure-aware trajectory modulation. Best in \textbf{\textcolor{red}{red}}, second in \textcolor{blue}{\underline{blue}}.}
\label{tab:ablation_components}
\renewcommand{\arraystretch}{1.15}
\scriptsize
\setlength{\tabcolsep}{6pt}
\begin{tabular}{ccc|cccc|c}
\toprule
\rowcolor{blue!10}
$x_0$+$v$ & Sparse & STM & PSNR$\uparrow$ & SSIM$\uparrow$ & MS-SSIM$\uparrow$ & HFEN$\downarrow$ & Train Time$\downarrow$ \\
\midrule
\xmark & \xmark & \xmark & 19.42 & .4639 & .5087 & 0.0812 & $1\times$ \\
\cmark & \xmark & \xmark & 31.25 & .8073 & .9613 & 0.0230 & $1\times$ \\
\cmark & \cmark & \xmark & \textcolor{blue}{\underline{31.56}} & \textcolor{blue}{\underline{.8137}} & \textcolor{blue}{\underline{.9624}} & \textcolor{blue}{\underline{0.0222}} & $\sim\!0.1\times$ \\
\rowcolor{gray!15}
\cmark & \cmark & \cmark & \textbf{\textcolor{red}{31.82}} & \textbf{\textcolor{red}{.8232}} & \textbf{\textcolor{red}{.9644}} & \textbf{\textcolor{red}{0.0215}} & $\sim\!0.1\times$ \\
\bottomrule
\end{tabular}
\end{table}

\section{Conclusion}
We presented a sparse voxel-space diffusion framework for 3D medical image enhancement that combines $x_0$-prediction with velocity supervision, uniform timestep subsampling, and structure-aware trajectory modulation (STM). The key insight is that strong anatomical priors in the degraded input render dense noise schedules largely redundant for conditional enhancement. By exploiting this redundancy, our framework trains and samples on as few as five timesteps while STM adapts denoising behavior to local anatomical structure. Experiments across CT, PET, and MRI demonstrate up to $10\times$ training acceleration with consistent improvements in reconstruction quality. We believe this principle extends naturally to other conditional generative tasks in volumetric medical imaging.



\bibliographystyle{splncs04}
\bibliography{main}
\end{document}